\pdfoutput=1
\PassOptionsToPackage{backref=page}{hyperref}

    \PassOptionsToPackage{compress}{natbib}

\documentclass{article} %
\usepackage{iclr2024_conference,times}

\newif\ifarxiv
\arxivtrue

\newif\ifperfect
\perfectfalse %

\newif\ifcompact
\compacttrue %

\ifarxiv
    \iclrfinaltrue
\else
\fi

\usepackage{comment}
\usepackage[utf8]{inputenc} %
\usepackage[T1]{fontenc}    %
\usepackage{hyperref}       %
\usepackage{url}            %
\usepackage{booktabs}       %
\usepackage{amsfonts}       %
\usepackage{nicefrac}       %
\usepackage{microtype}      %
\usepackage{xcolor}         %
\usepackage{xspace}
\newcommand{\tworowbrace}{
    \multirow{2}{*}{$\left\{
    \begin{array}{l}
    \vphantom{1}\\
    \vphantom{1}
    \end{array}\right.$}
    \hspace{-2em}
} %

\renewcommand{\paragraph}[1]{\noindent \textbf{#1}}

\usepackage{spverbatim}
\usepackage{listings} 

\usepackage{sec/zhijing_package} %
\newcommand{\ourdata}{{\textsc{Corr2Cause}}\xspace}

\usepackage{todonotes}

\newcommand\rada[1]{\rada[]{RM: #1}}

\newif\ifaccuracy

\title{
Can Large Language Models Infer \\Causation from Correlation?

}

\author{
Zhijing Jin\textsuperscript{\rm 1,2,\thanks{
Equal contribution. \textsuperscript{$\dagger$}Equal supervision. \textsuperscript{$\ddagger$}Work originated as a Meta AI internship project involving Zhijing, Mona, and Spencer.
}~~,$\ddagger$}\quad
Jiarui Liu\textsuperscript{\rm 3,\samethanks}\quad
Zhiheng Lyu\textsuperscript{\rm 4}\quad
Spencer Poff\textsuperscript{\rm 5}\quad
\\
\textbf{
Mrinmaya Sachan\textsuperscript{\rm 2}\quad
Rada Mihalcea\textsuperscript{\rm 6}\quad
Mona Diab\textsuperscript{\rm 3,$\ddagger$,$\dagger$
}\quad
Bernhard Sch\"olkopf\textsuperscript{\rm 1,$\dagger$
}}
\\
\textsuperscript{1}Max Planck Institute for Intelligent Systems, Tübingen, Germany\quad \textsuperscript{2}ETH Zürich\quad 
\\
\textsuperscript{3}LTI, CMU\quad
\textsuperscript{4}University of Hong Kong\quad
\textsuperscript{5}Meta AI\quad
\textsuperscript{6}University of Michigan
\\
\texttt{jinzhi@ethz.ch} \quad \texttt{jiarui@cmu.edu} \quad \texttt{zhihenglyu.cs@gmail.com}
\\
}

\begin{document}

\maketitle
\setcounter{footnote}{0}

\begin{abstract}
Causal inference
is one of the hallmarks of human intelligence. While the field of Causal NLP has attracted much interest in the recent years, existing causal inference datasets in NLP primarily rely on discovering causality from empirical knowledge (e.g., commonsense knowledge). In this work, we propose the first benchmark dataset to test the 
pure causal inference skills of large language models (LLMs).
Specifically, we formulate a novel task \ourdata, which takes a set of correlational statements
and determines the causal relationship between the variables.
We curate a large-scale dataset of more than 200K samples, on which we evaluate seventeen existing LLMs. Through our experiments, we identify a key shortcoming of LLMs in terms of their causal inference skills, and show that these models achieve almost close to random performance on the task. This shortcoming is somewhat mitigated when we try to re-purpose LLMs for this skill via finetuning, but we find that these models still fail to generalize -- they can only perform causal inference in in-distribution settings when variable names and textual expressions used in the queries are similar to those in the training set, but fail in out-of-distribution settings generated by perturbing these queries.
\ourdata is a challenging task for LLMs, and can be helpful in guiding future research on improving LLMs' pure reasoning skills and generalizability.\footnote{
\ifarxiv
Our data is at  {\small \url{https://huggingface.co/datasets/causalnlp/corr2cause}}.
\\
Our code is at {\small \url{https://github.com/causalNLP/corr2cause}}.
\else
Our code and data have been uploaded to the submission system, and will be open-sourced upon acceptance.
\fi
}
\end{abstract}

\section{Introduction}

Causal inference, i.e., the ability to establish the correct causal relationships between variables or events, is fundamental to human intelligence. %
There are two distinct ways this causal inference capability can be acquired: one through empirical knowledge, e.g., we know from common sense that touching a hot stove will get us burned;
the other through \textit{pure causal reasoning}, 
as causality can be formally argued and reasoned about using known procedures and rules from causal inference \citep{spirtes2000causation,pearl2009causality,peters2017elements}.
One example is that we have the a priori knowledge that the correlation between A and B does not necessarily imply causality. This is a formal rule that holds true regardless of the realizations of the variables A and B.

With the rise of large language models (LLMs) \citep[][\textit{inter alia}]{radford2019language,devlin-etal-2019-bert,ouyang2022instructGPT,
zhang2022opt,openai2023gpt4}, a crucial research question is whether they can do causal reasoning well. 
Recent studies have pointed out that LLMs are ``causal parrots,'' which recite the causal knowledge in the training data \citep{zevcevic2023causal}. Moreover, the vast majority of studies frame causal reasoning as a skill to navigate around empirical knowledge \citep{gordon-etal-2012-semeval,sap2019atomic,Sap2019SocialIQA,qin-etal-2019-counterfactual,bhagavatula2020abductive}, and also
treat LLMs as a knowledge base when evaluating its causal skills \citep{kiciman2023causal,tu2023causal,xie2023echo}.
However, all the above lines of research frame causality as empirical knowledge, thus relying heavily on the quality and the coverage of the training data,  overlooking the great potential of the formal causal reasoning skills 
to process correlational information to causal conclusions.

\begin{figure}[t]
\centering
    \includegraphics[width=\textwidth]{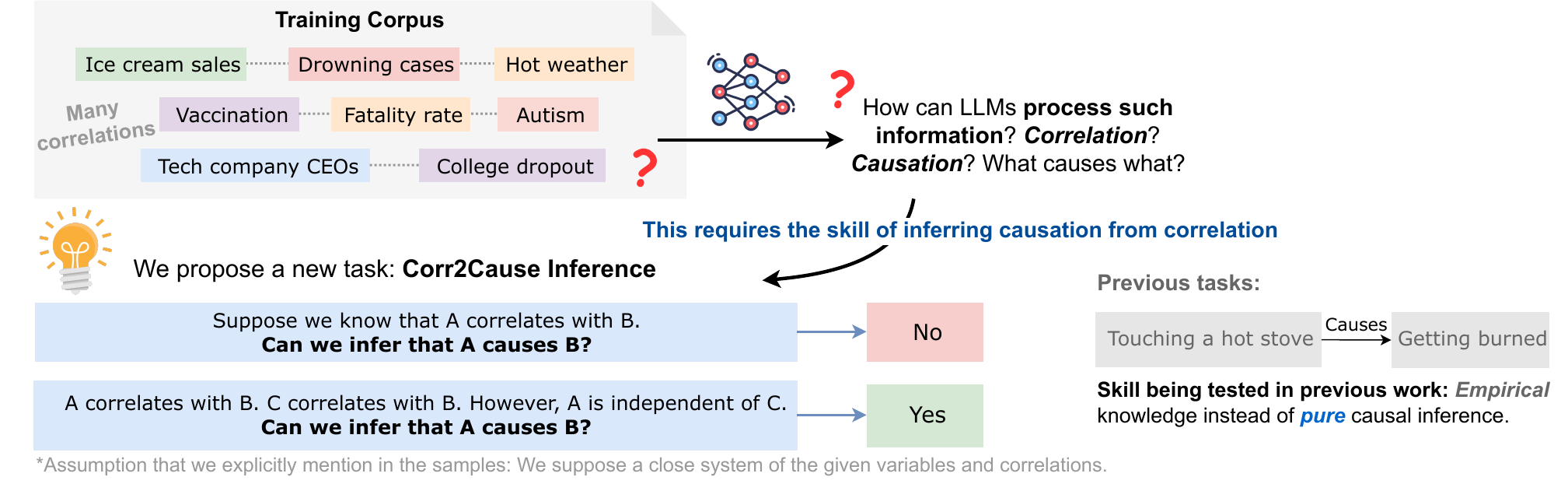}
    \vspace{-2em}
    \caption{Illustration of the motivation behind our task and dataset.}
    \label{fig:ex}
\vspace{-3mm}
\end{figure}

Drawing inspirations from technical studies on causal discovery \citep{spirtes2000causation,spirtes2016causal,clark2019review},
we formulate a novel task for NLP, \textit{correlation-to-causation inference }(\ourdata), which is an important skill for LLMs. Imagine the scenario in \cref{fig:ex}, where the training corpus does not tediously cover every causal relation, but more pervasively talk about correlations, such as which events tend to co-occur. Learning a good \ourdata skill can enable LLMs to draw causal relations behind the mere correlational information on the surface. For example, several decades ago, there might be an observation that female university students tend to perform better, but behind the correlational statistics is the causal graph that female students have to achieve extra good performance to get into universities as the first place.

To this end, we collect the \ourdata dataset, the first dataset %
to test the pure causal reasoning abilities of LLMs.
All the questions in this dataset are centered around testing
when it is valid or invalid to infer causation from correlation. To 
systematically compose this dataset, we ground our generalization process in the formal framework of causal discovery \citep{spirtes1993causation,spirtes2000causation,glymour2016causal,spirtes2016causal}, which provides rules about how to deduce 
causal relations
among variables 
given their statistical correlation in the observational data.
We generate more than 200K data points, 
and label a correlation-causation statement pair as valid if and only if there is a bijective mapping between the statistical correlation and the underlying causality.

Based on our \ourdata dataset with 200K samples, we investigate two main  research questions: 
(1) How well do existing LLMs perform on this task? 
(2) Can existing LLMs be re-trained or re-purposed on this task and obtain robust causal inference skills?
Through extensive experiments, we show empirically that none of the 17 existing LLMs we investigate perform well on this pure causal inference task. We also show that although LLMs can demonstrate better performance after being finetuned on the data, the %
causal inference skills attained by them are not robust. 
In summary, our contributions are as follows:
\begin{enumerate}
[topsep=0px,itemsep=0.5px]
    \item We propose the novel task of \ourdata, 
    to probe an aspect of LLM's reasoning
    ability, \textit{pure causal inference};
    \item We compose a dataset of over 200K samples, using insights from causal discovery;
    \item We evaluate the performance of 17 LLMs on our dataset, finding that all of them perform  poorly, close to the random baseline;
    \item We further explored whether LLMs can learn the skill through finetuning, and find that 
     LLMs fail to robustly acquire this skill in out-of-distribution settings. Finally, we suggest future work to explore more ways to enhance the pure causal inference skill in LLMs.
\end{enumerate}

\section{Preliminaries: Causal Inference} \label{sec:causality}

\subsection{Directed Graphical Causal Models (DGCMs)}\label{sec:graph_notations}
A directed graphical causal model (DGCM) is a commonly used representation to express the causal relations among a set of variables. 
Given a set of $N$ variables $\bm{X} = \{X_1, \dots, X_N \}$, we can encode the causal relations among them using a directed graph $\mathcal{G} := (\bm{X}, \bm{E})$, where $\bm{E}$ is the set of directed edges. Each edge $e_{i,j} \in \bm{E}$ represents a causal link $X_i \rightarrow X_j$, meaning that $X_i$ is a direct cause of $X_j$. 
In the context of this work, we take the common assumption of directed acyclic graphs (DAGs), which most causal discovery methods use \citep{clark2019review}, as graphs with cycles can make the causal discovery process arbitrarily hard.

Following the graph-theoretic terminology, we use an analogy of the ancestry tree to denote the relations between two variables.
For example, we call $X_i$ as a \textit{parent} of $X_j$ if there is a directed edge $X_i \rightarrow X_j$ in the graph, and, thus, $X_j$ is a \textit{child} of $X_i$. Similarly, we denote $X_i$ as an \textit{ancestor} of $X_j$ if there exists a directed path from $X_i$ to $X_j$, and, thus, $X_j$ is a \textit{descendent} of $X_i$. Note that a parent is a special case of an ancestor where the directed path has a length of 1. 

For convenience, we also introduce the notions for some special three-variable relations.
Given two variables $X_i$ and $X_j$, we call a third variable $X_k$ a \textit{confounder} (i.e., \textit{common cause}) if $X_k$ is a parent of both $X_i$ and $X_j$; a \textit{collider} (i.e., \textit{common effect}) if $X_k$ is a child of both $X_i$ and $X_j$; and a \textit{mediator} if $X_k$ is both a child of $X_i$, and a parent of $X_j$. 

\subsection{D-Separation and Markov Property}

\paragraph{D-Separation}
D-separation \citep{pearl1988probabilistic} is a fundamental concept in graphical models used to determine whether two sets of nodes $\bm{X}$ and $\bm{Y}$ in a DAG $\mathcal{G}$ are conditionally independent given a third set of nodes
$\bm{Z}$, where the three sets are disjoint.
We say that $\bm{X}$ and $\bm{Y}$ are d-separated by $\bm{Z}$ if all paths between any node in $\bm{X}$ and any node in $\bm{Y}$ are \textit{blocked} by the conditioning set $\bm{Z}$. A path between $\bm{X}$ and $\bm{Y}$ is blocked by $\bm{Z}$ if there exists a node $A\in \bm{Z}$ which satisfies one of the following conditions: $A$ is the 
parent node in a fork structure on the path (i.e., $\cdot \leftarrow A \rightarrow \cdot$); $A$ is the 
mediator node in a chain structure on the path (i.e., $\cdot \rightarrow A \rightarrow \cdot$); or in any collider structure on the path (i.e., $\cdot \rightarrow A \leftarrow \cdot$), $\bm{Z}$ does not contain $A$ or its descendants.

\paragraph{Markov Property}
The Markov property in a DAG $\mathcal{G}$ states that each node $X_i$ is conditionally independent of its non-descendants given its parents, namely
$X_i \perp\!\!\!\perp \NonDe(X_i) | \PA(X_i)$,
where $\NonDe(X_i)$ denotes the non-descendants of $X_i$ excluding itself, and $\PA(X_i)$ denotes the parents of $X_i$.
Using the Markov property, we can factorize the joint distribution of all the nodes in the graph into
$P(X_1, \dots, X_N) = \prod_{i=1}^N P(X_i | \bm{\mathrm{PA}}(X_i) )$. To infer the causal graph from probability distributions, a common assumption is faithfulness, namely the validity to infer all the d-separation sets in the graph from the independence relations in the probability distribution. In our work, we also take this broadly taken assumption which holds for most real-world scenarios.

\paragraph{Markov Equivalence of Graphs}
We denote two DAGs as Markov equivalent if they induce the same joint distribution $P(\bm{X})$. The set of DAGs that are Markov equivalent to each other is called a Markov equivalence class (MEC).
Causal graphs in the same MEC can be easily identified since they have the same skeleton (i.e., undirected edges) and V-structures (i.e., structures in the form of $A\rightarrow B \leftarrow C$ where $A$ and $C$ are not connected).

Obviously, there is a one-to-many mapping (i.e., surjection) between the causal graph and statistical distribution. 
Namely, each causal graph sufficiently determines a statistical distribution, but from a statistical distribution, we cannot necessarily induce a unique causal graph. This is why we say ``correlation does not necessarily mean causation''.

\subsection{Causal Discovery} 

Causal discovery aims to learn the causal relations by analyzing statistical
properties in the observational data \citep{spirtes1993causation,spirtes2000causation,glymour2016causal,spirtes2016causal,clark2019review}. It can be achieved through constraint-based methods \citep{spirtes2000causation}, score-based methods 
\citep{chickering2002optimal}, or other methods taking advantage of the functional causal models \citep{shimizu2006linear,hoyer2008nonlinear,zhang2009causality}.

To fit for the spirit of this paper to infer from correlation (expressed in natural language) to causation, we base our dataset design on the widely-used Peter-Clark (PC) algorithm \citep{spirtes2000causation}. 
The PC algorithm is based on the principles of conditional independence and the causal Markov assumption, which allows it to efficiently identify causal relationships among variables in a given dataset. 
The algorithm first starts with a fully connected undirected graph among all the variables. Then it removes the edge between two variables if there is an unconditional or conditional independence relationship between them. Afterwards, it orients the directed edges whenever there is a V-structure. And finally, it iteratively checks the direction of the other edges until the entire causal graph is consistent with all the statistical correlations.

\section{Dataset Construction}
\begin{figure}[t]
    \centering
    
    \includegraphics[width=0.85\columnwidth]{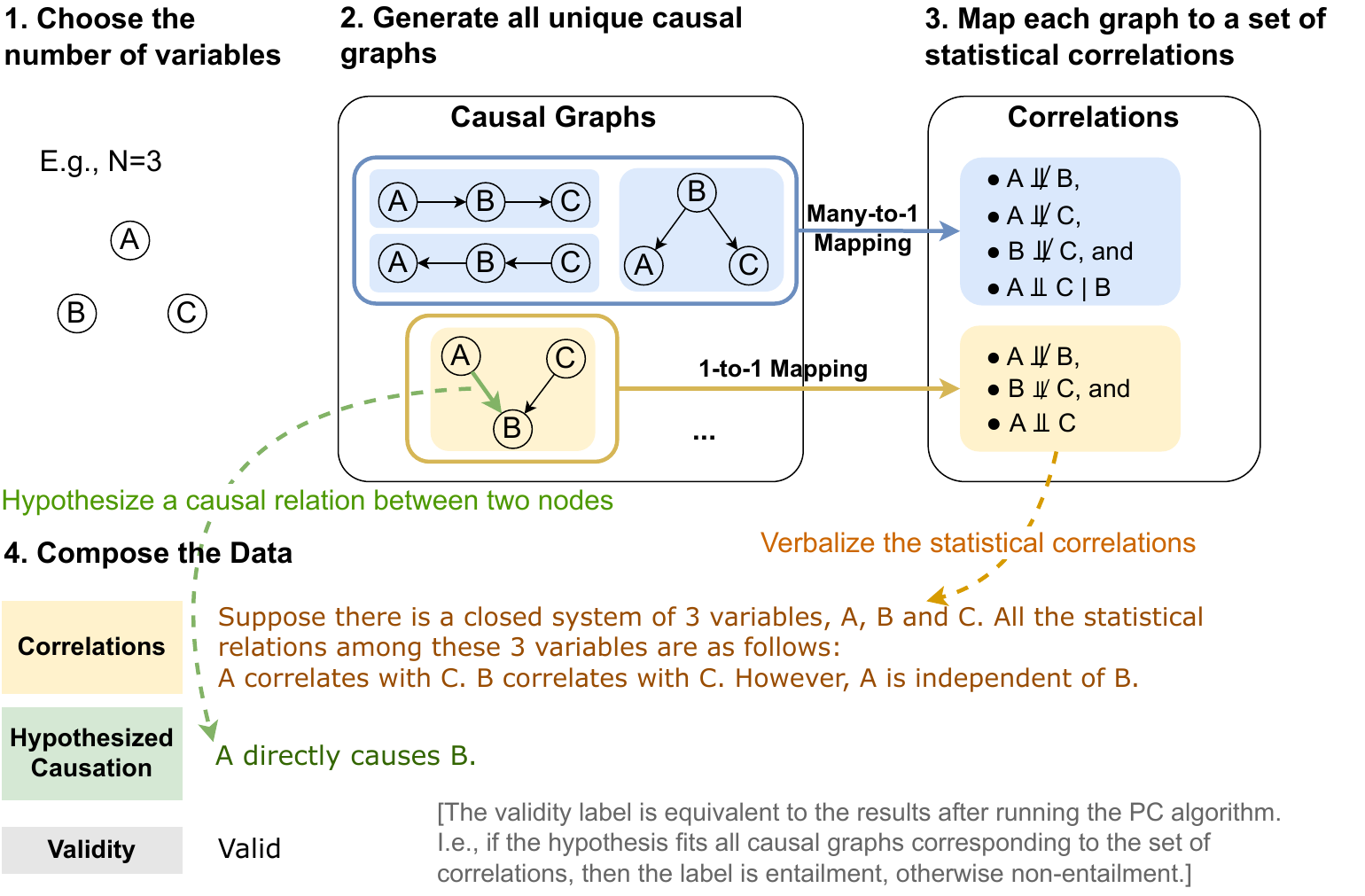}
    \caption{Pipeline of the data construction process.}
    \label{fig:construction}
\vspace{-3mm}
\end{figure}
We introduce the construction of our dataset in this section. We start with our task formulation for \ourdata, and then briefly give an overview of the data generation process, followed by detailed descriptions of each step. We conclude the section with the overall statistics of the dataset.

\subsection{Task Formulation}
Given a set of $N$ variables $\bm{X}=\{X_1, \dots, X_N\}$, we have a statement $\bm{s}$ about all the correlations among the variables, and a hypothesis $\bm{h}$ describing the causal relation $r$ between the pair of variables $X_i$ and $X_j$. The task is to learn a function $f: (\bm{s}, \bm{h}) \mapsto v$ which maps the correlation statement $\bm{s}$ and the causal relation hypothesis $\bm{h}$ to their validity $v \in \{0, 1\}$, which takes the value 0 if this inference is invalid, and the value 1 if this inference is valid.

\subsection{Overview of the Data Generation Process}
We base the construction our dataset on several concepts of causal inference, including the DGCM, d-separation, and MECs, as introduced in \cref{sec:causality}. 

As in the overview of our data generation process in \cref{fig:construction}, we first choose the number $N$ of variables (Step 1) and generate all the unique DGCMs with $N$ nodes (Step 2), which we will introduce in the \cref{sec:graph_generation}. Then we collect all the d-separation sets from these graphs to identify MECs (Step 3) in \cref{sec:dsep_generation}.
Then, in Step 4, we create the formal form of data in \cref{sec:form_generation}. For each correspondence of the MEC to causal graphs, we compose the correlation statement based on the statistical relations in the MEC, and  hypothesize a causal relation between two variables, and produce the validity $v=1$ if the hypothesis is a shared property of all causal graphs in the MEC, and $v=0$ if the hypothesis is not necessarily true for all the MEC graphs. Finally, we introduce the verbalization process in \cref{sec:verbalization}.

\subsection{Constructing the Graphs with Isomorphism Checks}\label{sec:graph_generation}

The first step of the data generation is to compose the causal graphs, as in Step 1 and 2 of \cref{fig:construction}.
For a set of $N$ variables $\bm{X} = \{X_1, \dots, X_N\}$, there are $N(N-1)$ possible directed edges, since each node can link to any node other than itself. 
To remove cycles in the graph, we make the nodes in topological order, which only allows edges $X_i \rightarrow X_j$, where $i<j$. We achieve this by limiting the adjacency matrix of the graph to only having non-zero values above the diagonal, resulting in $N(N-1)/2$ possible directed edges for the DAGs.

At the first glance, for $N$ nodes, there should be $2^{N(N-1)/2}$ possible DAGs (i.e., the power set of all edges).  However, there could be isomorphic graphs in this set. To avoid this,  we perform a graph isomorphism check \citep{mckay2014practical}, and reduce the set so that only unique DAGs are retained, and we show their statistics in \cref{tab:step1_causal_graph}. Although we can handle large graphs, we mostly focus on smaller graphs that can still lead to a reasonably sized dataset, so we empirically set $N=6$, but future work can use our open-sourced codes to extend to more nodes.

\begin{table}[t]
    \centering \small
    \begin{tabular}{clccccc}
    \toprule
    \# Nodes & \# Unique DAGs & \# Edges/DAG & \# MECs & \# DAGs/MEC \\ \midrule
    2 & 2 out of $2$ & 0.50 & 2 & 1.0 \\
    3 & 6 out of $2^{3}$ & 1.67 & 5 & 1.2 \\
    4 & 31 out of $2^{6}$ & 3.48 & 20 & 1.55 \\
    5 & 302 out of $2^{10}$ & 5.89 & 142 & 2.13 \\
    6 & 5,984 out of $2^{15}$ & 8.77 & 2,207  & 2.71 \\
    Total & 6,325 & 8.60 & 2,376  & 2.66 \\
    \bottomrule
    \end{tabular}
    \caption{Statistics about the source causal graphs in our dataset. Given the number of nodes, we report the number of unique DAGs, average number of edges per DAG, number of MECs, and average number of DAGs per MEC.}
    \label{tab:step1_causal_graph}
    \ifcompact \vspace{-3mm}\fi
\end{table}

\subsection{Programmatically Generating the D-Separation Sets}\label{sec:dsep_generation}
Based on the set of unique DAGs, we then programmatically generate the d-separation sets by graph theoretical conditions, as in Step 3 of \cref{fig:construction}. 
To realize this step, we code an efficient graph-theoretic algorithm to check for all the chain, fork, and collider structures to automatically identify the set of nodes that d-separate each pair of nodes.
Using the d-separation sets and the faithfulness assumption, we form the statistical correlations as follows. For each pair of nodes, they are conditionally independent given the variables in the d-separation set. If the d-separation set is empty, then the two nodes are unconditionally independent. If no d-separation set can be found for the two nodes, then they are directly correlated.

Moreover, using the d-separation sets, we are able to cluster causal graphs to MECs. We achieve it by tracing the mapping between the causal graphs and the set of statistical correlations, and backtracking the graphs with the same d-separation sets to group them in the same MEC. We show in \cref{tab:step1_causal_graph} that each MEC contains on average 2.66 DAGs.

\subsection{Composing the Hypotheses and Label}\label{sec:form_generation}

After generating the set of correlations based on the d-separation sets, we now generate the causal hypotheses. For the causal relation $r$, we focus on six common causal relations between two nodes introduced in \cref{sec:graph_notations}: Is-Parent, Is-Child,
Is-Ancestor (excluding the parents), Is-Descendant (excluding the children), 
Has-Confounder (i.e., there exists a confounder, or common cause, of the two nodes), and
Has-Collider (i.e., there exists a collider, or common effect, of the two nodes). 
In this way, the set of hypotheses contains 
all six meaningful causal relations between every pair of variables, resulting in a total size of
$6 \cdot N (N-1)/2 = 3N(N-1)$ hypotheses for a graph with $N$ variables.

To generate the ground-truth validity label, we start from the correlation sets in Step 3, then look up all the causal graphs in the same MEC corresponding to the given set of correlations, and check the necessity of the hypothesized causal relation. If the causal relationship proposed in the hypothesis is valid for all causal graphs within the MEC, then we generate the validity $v=1$; otherwise, we generate $v=0$. A special case of valid samples is that when the size of the MEC is 1, then there is a bijective mapping between the causal graph and the d-separation sets, so any hypothesis stating the causal properties of that unique causal graph is valid.

\subsection{Verbalizing into Language}\label{sec:verbalization}

Finally, as in the last step of \cref{fig:construction}, we convert all the information above to text data for our \ourdata task.
For the correlation statement, we verbalize
the set of correlations in Step 3 into a natural language statement $\bm{s}$.
When two variables cannot be d-separated, i.e., $A \not\perp\!\!\!\perp B$, then we describe them as ``$A$ correlates with $B$'' since they are directly correlated and cannot be independent by any condition. And if two variables have a valid d-separation set $\bm{C}$, then we describe them as ``$A$ is independent of $B$ given $\bm{C}$.'' In the special case when the d-separation set is empty, we directly say ``$A$ is independent of $B$.''
In addition, we disambiguate the setting by starting the correlation statement with the setup of a closed system of the given variables, and no hidden variables: ``Suppose there is a closed system of $N$ variables, A, B, \dots ~All the statistical relations among these $N$ variables are as follows:''.
Finally, to verbalize the hypothesis, we feed the causal relation triplet ($X_i$, $r$, $X_j$) into their hypothesis templates in \cref{tab:template}. For example, we turn the triplet ($A, \text{Is-Parent}, B$) into ``$A$ directly causes $B$'', as in the example of \cref{fig:construction}.
\begin{table}[t]
    \centering \small
    \begin{tabular}{ll}
    \toprule
Causal Relation & Hypothesis Template \\\midrule
Is-Parent & \texttt{\{Var i\}} directly causes \texttt{\{Var j\}}. \\
Is-Ancestor & \texttt{\{Var i\}} causes something else which causes \texttt{\{Var j\}}. \\
Is-Child & \texttt{\{Var j\}} directly causes \texttt{\{Var i\}}. \\
Is-Descendant & \texttt{\{Var j\}} is a cause for \texttt{\{Var i\}}, but not a direct one. \\
Has-Collider & There exists at least one collider (i.e., common effect) of \texttt{\{Var i\}} and \texttt{\{Var j\}}. \\
Has-Confounder & There exists at least one confounder (i.e., common cause) of \texttt{\{Var i\}} and \texttt{\{Var j\}}. \\
\bottomrule
    \end{tabular}
    \caption{Templates for each causal relation in the hypothesis. We use \texttt{\{Var i\}} and \texttt{\{Var j\}} as placeholders for the two variables.}
    \label{tab:template}
\vspace{-5mm}
\end{table}

\subsection{Statistics of the Resulting Data}\label{sec:stats} 

We show the statistics of our \ourdata{} dataset in \cref{tab:stats}. 
Overall, our dataset contains 207,972 samples, where 18.57\% of the samples have the positive label (i.e., with validity=1). 
The average length of the premise is 424.11 tokens, and hypothesis 10.83 tokens. We split the data into 205,734 training samples, 1,076 development and 1,162 test samples.\footnote{\textit{Note for our dataset v2.0:} We notice that our original data (v1.0) has duplication due to symmetric relations and verbalizations of the hypothesis. E.g., Is-Parent(A, B) has the exact hypothesis verbalization as Is-Child(B, A). Hence, for our data v2.0, we perform a careful de-duplication, and update the data statistics in \cref{tab:stats}. See more version comparison details in \cref{appd:version}. Note that, due to the symmetry, the current version is a random sample half of the size of the original version, so the modeling results in the experiment section roughly hold.} Since the main purpose of the dataset is
to benchmark the performance of LLMs, we prioritize 
the test and development sets to have a comprehensive coverage over all sizes of graphs. Specifically, we iterate through the subset of our data for each $N$, and split it entirely for only the
test and development sets if the data is less than 1K, which is the case for $N=2$ and $3$. For the other subsets that are larger, we randomly sample up to 1K or 10\% of the data, whichever is smaller, to the test and development sets. We set the cap to be 1K in order to form a reasonable computation budget, since many LLMs are expensive to query in the inference mode. Aside from the test and valid sets, all the rest of the data goes into the training set.

\begin{table}[h]
    \centering\small
    \begin{tabular}{lcccccccccccc}
    \toprule
& \multirow{2}{*}{Overall} & \multicolumn{5}{c}{Statistics by the Number of Nodes $N$} \\
\cline{3-7}
&  & $N=2$ & $N=3$ & $N=4$ & $N=5$ & $N=6$ \\ \midrule
\# Samples & 207,972 & 12 & 90 & 720 & 8,520 & 198,630  \\
\quad \# Test  & 1,162 & 6 & 48 & 72 & 514 & 522  \\
\quad \# Dev & 1,076 & 6 & 42 & 72 & 482 & 474  \\
\quad \# Train & 205,734 & 0 & 0 & 576 & 7,524 & 197,634  \\
\# Tokens/Premise & 424.11 & 31.5 & 52.0 & 104.0 & 212.61 & 434.54  \\
\# Tokens/Hypothesis & 10.83 & 10.83 & 10.83 & 10.83 & 10.83 & 10.83  \\
\% Positive Labels & 18.57 & 0.00 & 3.33 & 7.50 & 13.01 & 18.85  \\
Vocab Size   & 65 & 49 & 53 & 55 & 57 & 61 \\
\bottomrule
    \end{tabular}
    \caption{Statistics of our \ourdata dataset, and by subsets. We report the total number of samples (\# Samples); splits of the test (\# Test), developement (\# Dev) and training sets (\# Train); number of tokens per premise (\# Tokens/Premise) and hypothesis (\# Tokens/Hypothesis); percentage of the positive labels (\% Positive Labels), and vocabulary size by the number of unique tokens (Vocab Size). Note that the number of unique graphs and MECs are in \cref{tab:step1_causal_graph}.}
    \label{tab:stats}
\end{table}

\vspace{-5mm}
\section{Experiments}

\subsection{Experimental Setup}
\accuracytrue

\ifaccuracy
\begin{table}[t]
    \centering \small
    \begin{tabular}{lccccc}
    \toprule
    & F1 & Precision & Recall
    & Accuracy\\ \midrule
    \multicolumn{5}{l}{\textit{\textbf{Random Baselines}}} \\
\quad    Always Majority & 0.0 & 0.0 & 0.0 & 84.77\\ 
\quad    Random (Proportional) & 13.5 & 12.53 & 14.62 & 71.46\\
\quad    Random (Uniform) & \ul{20.38} & 15.11 & 31.29 & 62.78\\
    \midrule
    \textit{\textbf{BERT-Based Models}} \\
\quad    BERT MNLI & 2.82 & 7.23 & 1.75 & 81.61\\
\quad    RoBERTa MNLI & 22.79 & 34.73 & 16.96 & 82.50\\
\quad    DeBERTa MNLI & 14.52 & 14.71 & 14.33 & 74.31\\ 
\quad    DistilBERT MNLI & 20.70 & 24.12 & 18.13 & 78.85\\ 
\quad    DistilBART MNLI & 26.74 & 15.92 & 83.63 & 30.23\\ 
\quad    BART %
    MNLI & \textbf{\ul{33.38}} & 31.59 & 35.38 & 78.50\\  \midrule
    \textit{\textbf{LLaMa-Based Models}} \\
\quad LLaMa-7B & 26.81 & 15.50 & 99.42 & 17.36\\ 
\quad Alpaca-7B & \ul{27.37} & 15.93 & 97.37 & 21.33\\
\midrule
    \textit{\textbf{GPT-Based Models}} \\
\quad GPT-3 Ada & 0.00 & 0.00 & 0.00 & 84.77\\ 
\quad GPT-3 Babbage & 27.45 & 15.96 & 97.95 & 21.15\\ 
\quad GPT-3 Curie & 26.43 & 15.23 & 100.00 & 15.23\\ 
\quad GPT-3 Davinci & 27.82 & 16.57 & 86.55 & 31.61\\ 
\quad GPT-3 Instruct (text-davinci-001) & 17.99 & 11.84 & 37.43 & 48.04\\ 
\quad GPT-3 Instruct (text-davinci-002) & 21.87 & 13.46 & 58.19 & 36.69\\ 
\quad GPT-3 Instruct (text-davinci-003) & 15.72 & 13.4 & 19.01 & 68.97\\
\quad     GPT-3.5 & 21.69 & 17.79 & 27.78 & 69.46\\ 
\quad GPT-4 & \ul{29.08} & 20.92 & 47.66 & 64.60\\ 
    \bottomrule
    \end{tabular}
    \caption{Overall performance. We report F1 (main metric), precision, recall and accuracy. For the main metric, F1 score, we use the \textbf{bold} font to highlight the overall best performance, and \ul{underline} to highlight the best performance within each category of models. 
    }
    \label{tab:res}
\vspace{-3mm}
\end{table}

\else

\begin{table}[t]
    \centering \small
    \begin{tabular}{lcccc}
    \toprule
    & F1 & Precision & Recall
    \\ \midrule
    \multicolumn{4}{l}{\textit{\textbf{Random Baselines}}} \\
\quad    Always Majority & 0.0 & 0.0 & 0.0 \\ 
\quad    Random (Proportional) & 13.5 & 12.53 & 14.62 \\
\quad    Random (Uniform) & \ul{20.38} & 15.11 & 31.29 \\
    \midrule
    \textit{\textbf{BERT-Based Models}} \\
\quad    BERT MNLI & 2.82 & 7.23 & 1.75 \\
\quad    RoBERTa MNLI & 22.79 & 34.73 & 16.96 \\
\quad    DeBERTa MNLI & 14.52 & 14.71 & 14.33 \\ 
\quad    DistilBERT MNLI & 20.70 & 24.12 & 18.13 \\ 
\quad    DistilBART MNLI & 26.74 & 15.92 & 83.63 \\ 
\quad    BART %
    MNLI & \textbf{\ul{33.38}} & 31.59 & 35.38 \\  \midrule
    \textit{\textbf{LLaMa-Based Models}} \\
\quad LLaMa-7B & 26.81 & 15.50 & 99.42 \\ 
\quad Alpaca-7B & \ul{27.37} & 15.93 & 97.37 \\
\midrule
    \textit{\textbf{GPT-Based Models}} \\
\quad GPT-3 Ada & 0.00 & 0.00 & 0.00 \\ 
\quad GPT-3 Babbage & 27.45 & 15.96 & 97.95 \\ 
\quad GPT-3 Curie & 26.43 & 15.23 & 100.00\\ 
\quad GPT-3 Davinci & 27.82 & 16.57 & 86.55 \\ 
\quad GPT-3 Instruct (text-davinci-001) & 17.99 & 11.84 & 37.43 \\ 
\quad GPT-3 Instruct (text-davinci-002) & 21.87 & 13.46 & 58.19 \\ 
\quad GPT-3 Instruct (text-davinci-003) & 15.72 & 13.4 & 19.01\\
\quad     GPT-3.5 & 21.69 & 17.79 & 27.78 \\ 
\quad GPT-4 & \ul{29.08} & 20.92 & 47.66\\ 
    \bottomrule
    \end{tabular}
    \caption{Overall performance. We report F1 (main metric), precision, recall and accuracy. For the main metric, F1 score, we use the \textbf{bold} font to highlight the overall best performance, and \ul{underline} to highlight the best performance within each category of models. 
    }
    \label{tab:res}
\vspace{-3mm}
\end{table}

\fi

We set up a diverse list of LLMs for the experiments on our \ourdata dataset. 
To \textit{test existing LLMs}, we first include six commonly used BERT-based NLI models in the transformers library \citep{wolf2019transformers}: 
BERT \citep{devlin-etal-2019-bert}, RoBERTa~\citep{liu2019roberta}, BART \citep{lewis-etal-2020-bart}, DeBERTa~\citep{he2021deberta}, DistilBERT~\citep{sanh2019distilbert}, and DistilBART \citep{shleifer2020pretrained}. Apart from these BERT-based NLI models, we also evaluate the general-purpose autoregressive LLMs based on GPT \citep{radford2019language}: GPT-3 Ada, Babbage, Curie, Davinci \citep{gpt3}; its instruction-tuned versions \citep{ouyang2022instructGPT}, text-davinci-001, text-davinci-002, and text-davinci-003; and GPT-3.5 (i.e., ChatGPT), and the latest GPT-4 \citep{openai2023gpt4} by April 2023, using the OpenAI API
({\small \url{https://openai.com/api/}})
with temperature 0. We also evaluate the recent, more efficient models, LLaMa \citep{touvron2023llama} and Alpaca \citep{taori2023alpaca}.

When inspecting the behavior of \textit{finetuned models}, 
we adopt a large set of models, including GPT-based models (GPT-3 Ada, Babbage, Curie, and Davinci) using the OpenAI finetuning API for classification at {\small \url{https://platform.openai.com/docs/guides/fine-tuning}},
open-sourced decoder-only models (GPT2, GPT2-Large, GPT2-XL, LLaMA-7B, and LLaMA2-7B), BERT-based models from scratch (BERT-Base, BERT-Large, RoBERTa-Base, and RoBERTa-Large), and BERT-Based NLI models (BERT-Base MNLI, BERT-Large MNLI, RoBERTa-Base MNLI, and RoBERTa-Large MNLI) using the transformers library \citep{wolf2019transformers}. See training details
in \cref{appd:implementation}.

For the \textit{random baselines}, we provide ``always majority'' to predict the majority class 100\% of the time, ``random (uniform)'' to uniformly sample a label (i.e., 50\% for each), and ``random (proportional)'' to sample a label from a Bernouli distribution proportional to the development set label distribution.

\subsection{The \ourdata Skill in Existing LLMs}\label{sec:0shot}

We show the performance of seventeen LLMs in \cref{tab:res}. We can see that pure causal inference is a very challenging task across all existing LLMs.
Among all the LLMs, the best performance is 33.38\% F1 by BART MNLI, which is even higher than the latest GPT-based model, GPT-4.
Notably, many models are worse than random guess, which means that they totally fail at this pure causal inference task.
The observation still holds for few-shot chain-of-thought prompts tested in \cref{appd:optimization}.

\subsection{Finetuned Performance}
Next, we address the question: \textit{Can we re-purpose LLMs to learn this task?}
The experimental results in \cref{tab:finetune} of 17 models finetuned on our \ourdata seem very strong at first sight. Most models see a substantial increase, among which the finetuned BERT-based NLI models demonstrate the strongest performance.
The best-performing one, RoBERTa-Large MNLI, achieves 94.74\% F1 score on this task, as well as very high precision, recall and accuracy scores.

\begin{table}[t]
    \centering \small
    \begin{subtable}[t]{0.66\textwidth}
        \centering
    \begin{tabular}{lccccc}
    \toprule
    & F1 & Precison & Recall
    & Accuracy  \\ \midrule
    \multicolumn{5}{l}{\textit{\textbf{Finetuned GPT-Based Models Using OpenAI API}}} \\

GPT-3 Ada & 79.85 & 70.47 & 92.11 & 92.92 \\ 
GPT-3 Babbage & 78.19 & 69.98 & 88.60 & 92.48 \\ 
GPT-3 Curie & 81.23 & 75.00 & 88.60 & 93.77 \\ 
GPT-3 Davinci & \ul{85.52} & 80.26 & 91.52 & 95.28 \\ \midrule
    \multicolumn{5}{l}{\textit{\textbf{Finetuned Open-Sourced Decoder-Only Models}}} \\
GPT2 & 89.18 & 88.03 & 90.35 & 96.66 \\
GPT2-Large & 94.29 & 92.18 & 96.49 & 98.22 \\
GPT2-XL & \ul{94.30} & 91.94 & 96.78 & 98.22 \\
LLaMA-7B &  91.98 & 88.62 & 95.61 & 97.46 \\
LLaMA2-7B & 92.92 & 90.11 & 95.91 & 97.77 \\ \midrule
    \multicolumn{5}{l}{\textit{\textbf{Finetuned BERT-Based Models}}} \\
    BERT-Base & 69.29 & 54.42 & 95.32 & 87.13 \\ 
BERT-Large & 85.26 & 77.51 & 94.74 & 95.01 \\ 
RoBERTa-Base & 87.60 & 78.47 & 99.12 & 95.73 \\ 
RoBERTa-Large & \ul{89.10} & 82.54 & 96.78 & 96.39 \\
    
\midrule
    \multicolumn{5}{l}{\textit{\textbf{Finetuned BERT-Based NLI Models}}} \\
    BERT-Base MNLI & 89.88 & 85.49 & 94.74 & 86.51 \\ 
    BERT-Large MNLI & 90.19 & 84.44 & 96.78 & 96.79 \\
    RoBERTa-Base MNLI & 94.27 & 90.35 & 98.54 & 98.17 \\
    RoBERTa-Large MNLI & \textbf{\ul{94.74}} & 92.24 & 97.37 & 98.35 \\

    \bottomrule
    \end{tabular}
    \caption{Performance of finetuned models on the original test set.
    }
    \label{tab:finetune}
\end{subtable}%
    \hfill
    \begin{subtable}[t]{0.3\textwidth}
        \centering
    \begin{tabular}{ccc}
    \toprule
    F1 (Paraph.) & F1 (Var. Ref.)  \\ \midrule
 \\
61.73 & 41.57 \\ 
62.34 & 43.28 \\ 
64.93 & 45.32 \\ 
65.01 & 46.96 \\ \midrule
\\
56.76 &   31.70 \\
55.95 &     31.99 \\
60.32 &    43.95 \\
56.41 &  53.92 \\
52.24 &     49.47 \\ \midrule
\\
61.13 & 35.20 \\
63.64 & 38.54 \\
65.58 & 53.12 \\
65.05 & 60.20 \\
    
\midrule
\\
65.56 & 31.50 \\
67.24 & 52.04 \\
57.42 & 62.83 \\
55.45 & 67.87 \\
    \bottomrule
    \end{tabular}
    \caption{F1 scores of finetuned models on the perturbed test sets by paraphrasing (Paraph.) and variable refactorization (Var. Ref.).
    }
    \label{tab:goodhart}
\end{subtable}%
\vspace{-3mm}
\caption{Performance of finetuned models on the original test set and perturbed test sets.}
\vspace{-1mm}
\end{table}

\begin{table}[t]
    \centering \small
    
    \setlength{\tabcolsep}{5pt}
    \begin{subtable}[t]{0.6\textwidth}
    \begin{tabular}{lccccc}
\toprule
Relation Type & F1 & Precision & Recall & Accuracy \\ \midrule
Is-Parent &  96.18 &  95.45 &  96.92 & 98.67 \\
Is-Ancestor &  93.94 &  93.94 &  93.94 & 98.93 \\
Is-Child &  95.73 &  94.92 &  96.56 & 98.67 \\
Is-Descendant &  96.55 &  93.33 &  100 & 99.47 \\
Has-Collider &  92.19 &  87.41 &  97.52 &  94.64 \\
Has-Confounder &  98.67 &  97.37 &  100 & 99.73 \\

\bottomrule
    \end{tabular}
    \caption{Fine-grained performance of RoBERTa-Large by causal relation type on the original test set.
    }
    \label{tab:finetune_by_class}
\end{subtable}%
    \hfill
    \begin{subtable}[t]{0.38\textwidth}
        \centering
        
    \setlength{\tabcolsep}{5pt}
    \begin{tabular}{cccccc}
    \toprule

F1 & Precision & Recall & Accuracy \\ \midrule

74.80 & 79.31 & 70.77 &	91.73 \\
45.45 & 90.91 & 30.30 &	93.60 \\
73.39 & 78.43 & 68.97 &	92.27 \\
29.41 & 83.33 & 17.86 &	93.60 \\
70.70 & 75.00 & 66.90 &	82.04 \\
70.42 & 73.53 & 67.57 &	94.37 \\

    \bottomrule
    \end{tabular}
    \caption{Its fine-grained performance by relation type after variable refactorization.
    }
    \label{tab:perturb_by_class}
\end{subtable}%
\vspace{-2mm}
\caption{Fine-grained analysis of the best-performing model, RoBERTa-Large MNLI.}
\vspace{-2mm}
\end{table}

\subsection{Fine-Grained Performance by Causal Relation}
In addition to the overall results mentioned above, we conduct a fine-grained analysis to check the performance of the strongest finetuned model, RoBERTa-Large MNLI, by our six causal relation types. As in \cref{tab:finetune_by_class}, the model is very good at judging relations such as Is-Parent, Is-Descendant and Has-Confounder, all with more than 96\% F1 scores, whereas it is several points weaker on the Has-Collider relations. This could be due to that the collider relation is the most special type, requiring identification of the V-structure based on both the unconditional independence based on the two variables only and correlations whenever conditioned on a common descendant. We also conduct error analysis for non-finetuned models in \cref{appd:error}.

\subsection{Robustness Analysis
}

Looking at the very high performance of the finetuned models, we raise the next question: \textit{Did the models really robustly learn the causal inference skills?} 

\paragraph{Two Robustness Tests}
We design two simple robustness tests: (1) paraphrasing, and (2) variable refactorization. For (1) paraphrasing, we simply paraphrase the hypothesis by changing the text template for each causal relation to some semantically-equivalent alternatives in \cref{appd:templates}. For (2) variable refactorization, we reverse the alphabet of the variable names, namely flipping A, B, C, to Z, Y, X and so on.
The inspiration behind the two robustness tests comes from the spurious correlation analysis described in \cref{appd:spurious}.

Specifically, we adopt the common setup of text adversarial attack \citep{morris-etal-2020-textattack,jin2020bert} to preserve the training set and keep the same saved models, but run the inference on the perturbed test set.
In this way, we separate the possibilities of the models only overfitting on the training data vs. mastering the reasoning skills.

\paragraph{Results after Perturbation}
We can see from \cref{tab:goodhart} that all the models drop drastically, by up to 39.29 on the paraphrased test set, and up to 62.30 after variable refactorization. 
The best-performing model, RoBERTa-Large MNLI, is especially sensitive towards paraphrasing, demonstrating the most drop among all models; however, it is the most robust against the variable refactorization, maintaining a high F1 score of 67.87. We conduct fine-grained  analysis for RoBERTa-Large MNLI under perturbation in \cref{tab:perturb_by_class}. We can see the the main source of the performance drop of the model comes from the two classes, Is-Ancestor (decreasing to 45.45\%) and Is-Descendant (decreasing to 29.41\%), while the other classes stay relatively robust, keeping their F1 scores over 70\%.

From this analysis, we make the following suggestions to future studies testing this \ourdata skill of LLMs. First, it is safe to use it as a test set to benchmark existing LLMs' performance, since the data we generate is out-of-distribution from the training data of the current LLMs. Then, when testing finetuned models, it is very important to accompany adversarial attack together with the i.i.d. test set. We open-source our perturbed test sets for future work to test the generalizability skill.

\ifperfect
\subsection{Effect of In-Context Learning}
\subsection{Adding Signals to the context}
\subsection{Example Reasoning}

\fi
\vspace{-1mm}
\subsection{Extension to Natural Stories}

We envision our \ourdata dataset to be a foundation for future extensions to various settings, such as instantiating the variables with actual phenomena and situating the story in a more natural setting.
For example, the \textit{correlation does not imply causation} rule can be instantiated with the ice cream sales and swimming pool attendance as the two variables, and argue that
ice cream sales does not necessarily affect swimming pool attendance, because their correlation could be due to a third variable, such as hot weather. We provide a case study for how to instantiate the symbolic expressions in our dataset to more natural stories, and find that LLMs such as GPT-4 can generate realistic, daily life stories that has foreseeably broad applications. See more details
in \cref{appd:story}.

\vspace{-1mm}
\section{Related Work}
\paragraph{Existing Causal Reasoning Tasks}
A large body of existing research of causal reasoning in NLP focuses on leveraging empirical knowledge to do tasks such as inferring the cause and effect of why an agent perform certain tasks \citep{sap2019atomic}, the motivation and emotional reaction in a social context \citep{Sap2019SocialIQA}, how people achieve a given goal with a set of concrete steps \citep{zhang-etal-2020-reasoning}, the development of a story given a different beginning \citep{qin-etal-2019-counterfactual}, and how in general LLMs serve as a knowledge base of cause and effect \citep{willig2023probing,kiciman2023causal}.
In contrast, our \ourdata task focuses on the pure causal inference skill of models, which is a knowledge-dependent reasoning skill based on formally correct rules from causal inference.

\paragraph{Existing Logical and Inference Tasks}
Another related area of literature is logical and inference tasks, of which a well-established one is natural language inference (NLI), to identify the semantic relationship between a pair of
sentences \citep{maccartney-manning-2008-modeling,bowman-etal-2015-large}. NLI datasets mainly focus on the set and paraphrase relations. For example, ``a group of boys are playing football'' can entail ``some guys are playing football,'' where ``boys'' are a sub-concept of ``guys,'' and ``a group of'' and ``some'' are paraphrases. 
Recently, there have been increasing efforts to extend the inference task to various logical inference skills such as deductive logic and propaganda techniques \citep{jin-etal-2022-logical,alhindi-etal-2022-multitask}. Our \ourdata dataset is the first dataset testing the correlation-to-causation inference skill, which is unique of its type.

\vspace{-1mm}
\section{Conclusion}
In this work, we introduced a novel task, \ourdata, to infer causation from correlation, and collected a large-scale dataset of over 200K samples.
We evaluated an extensive list of LLMs on this new task, and showed that off-the-shelf LLMs perform poorly on this task. We also show that it is possible to re-purpose LLMs on this task by finetuning, but future work needs to be aware of the out-of-distribution generalization problem. To avoid the Goodhart's law, we recommend using this dataset to benchmark the pure causal inference skills for LLMs that have not seen this dataset. 
Given the limited reasoning abilities of current LLMs, and the difficulty of separating actual reasoning from training-corpus-derived knowledge, it is imperative that our community focus on work aiming to accurately disentangle and measure both abilities. We believe the present work is a first such step.

\section*{Limitations and Future Work}
We identify several limitations of this work and open future directions:
First, in the context of this work, we limit the causal graphs to two to six nodes, but future work can feel free to explore larger graphs.
Another aspect is that we do not assume hidden confounders in this inference problem, so we welcome future work to generate an even more challenging dataset to infer the existence of hidden confounders, analogous to the causal discovery algorithm of fast causal inference (FCI)  \citep{spirtes2000causation}. And also in general, explorations of other causal discovery algorithms are welcomed too.
Finally, a lot of motivation behind proposing this task is inspired by the problem of invalid reasoning patterns in our daily reasoning \citep{jin-etal-2022-logical}, which could fertilize the ground for more pervasive spread of fake news. We believe false causal inference is a prevalent type of fallacious beliefs, and welcome future work to connect the idea of this benchmark to more real-world false beliefs based on confusing correlation with causation.

\section*{Acknowledgment}
We thank Riley Goodside for valuable suggestions to improve our prompts to LLMs. We thank Luigi Gresele and Amir Hossein Karimi for their suggestions to help us
improve the formulation of our causal discovery questions.

This material is based in part upon work supported by the German Federal Ministry of Education and Research (BMBF): Tübingen AI Center, FKZ: 01IS18039B; by the Machine Learning Cluster of Excellence, EXC number 2064/1 – Project number 390727645; by 
a National Science Foundation award (\#2306372); by a Swiss National Science Foundation award (\#201009) and a Responsible AI grant by the Haslerstiftung.
Zhijing Jin is supported by PhD fellowships from the Future of Life Institute and Open Philanthropy. We also thank OpenAI for granting Zhijing quota to their API of GPT series through the Researcher Access Program.

\bibliography{refs,sec/refs_causality,sec/refs_cogsci,sec/refs_zhijing}

\begin{thebibliography}{44}
\providecommand{\natexlab}[1]{#1}
\providecommand{\url}[1]{\texttt{#1}}
\expandafter\ifx\csname urlstyle\endcsname\relax
  \providecommand{\doi}[1]{doi: #1}\else
  \providecommand{\doi}{doi: \begingroup \urlstyle{rm}\Url}\fi

\bibitem[Alhindi et~al.(2022)Alhindi, Chakrabarty, Musi, and
  Muresan]{alhindi-etal-2022-multitask}
Tariq Alhindi, Tuhin Chakrabarty, Elena Musi, and Smaranda Muresan.
\newblock Multitask instruction-based prompting for fallacy recognition.
\newblock In \emph{Proceedings of the 2022 Conference on Empirical Methods in
  Natural Language Processing}, pp.\  8172--8187, Abu Dhabi, United Arab
  Emirates, December 2022. Association for Computational Linguistics.
\newblock URL \url{https://aclanthology.org/2022.emnlp-main.560}.

\bibitem[Bhagavatula et~al.(2020)Bhagavatula, Bras, Malaviya, Sakaguchi,
  Holtzman, Rashkin, Downey, Yih, and Choi]{bhagavatula2020abductive}
Chandra Bhagavatula, Ronan~Le Bras, Chaitanya Malaviya, Keisuke Sakaguchi, Ari
  Holtzman, Hannah Rashkin, Doug Downey, Wen{-}tau Yih, and Yejin Choi.
\newblock Abductive commonsense reasoning.
\newblock In \emph{8th International Conference on Learning Representations,
  {ICLR} 2020, Addis Ababa, Ethiopia, April 26-30, 2020}. OpenReview.net, 2020.
\newblock URL \url{https://openreview.net/forum?id=Byg1v1HKDB}.

\bibitem[Bowman et~al.(2015)Bowman, Angeli, Potts, and
  Manning]{bowman-etal-2015-large}
Samuel~R. Bowman, Gabor Angeli, Christopher Potts, and Christopher~D. Manning.
\newblock A large annotated corpus for learning natural language inference.
\newblock In \emph{Proceedings of the 2015 Conference on Empirical Methods in
  Natural Language Processing}, pp.\  632--642, Lisbon, Portugal, September
  2015. Association for Computational Linguistics.
\newblock \doi{10.18653/v1/D15-1075}.
\newblock URL \url{https://aclanthology.org/D15-1075}.

\bibitem[Brown et~al.(2020)Brown, Mann, Ryder, Subbiah, Kaplan, Dhariwal,
  Neelakantan, Shyam, Sastry, Askell, Agarwal, Herbert-Voss, Krueger, Henighan,
  Child, Ramesh, Ziegler, Wu, Winter, Hesse, Chen, Sigler, Litwin, Gray, Chess,
  Clark, Berner, McCandlish, Radford, Sutskever, and Amodei]{gpt3}
Tom Brown, Benjamin Mann, Nick Ryder, Melanie Subbiah, Jared~D Kaplan, Prafulla
  Dhariwal, Arvind Neelakantan, Pranav Shyam, Girish Sastry, Amanda Askell,
  Sandhini Agarwal, Ariel Herbert-Voss, Gretchen Krueger, Tom Henighan, Rewon
  Child, Aditya Ramesh, Daniel Ziegler, Jeffrey Wu, Clemens Winter, Chris
  Hesse, Mark Chen, Eric Sigler, Mateusz Litwin, Scott Gray, Benjamin Chess,
  Jack Clark, Christopher Berner, Sam McCandlish, Alec Radford, Ilya Sutskever,
  and Dario Amodei.
\newblock Language models are few-shot learners.
\newblock In H.~Larochelle, M.~Ranzato, R.~Hadsell, M.~F. Balcan, and H.~Lin
  (eds.), \emph{Advances in Neural Information Processing Systems}, volume~33,
  pp.\  1877--1901. Curran Associates, Inc., 2020.
\newblock URL
  \url{https://proceedings.neurips.cc/paper/2020/file/1457c0d6bfcb4967418bfb8ac142f64a-Paper.pdf}.

\bibitem[Chickering(2002)]{chickering2002optimal}
David~Maxwell Chickering.
\newblock Optimal structure identification with greedy search.
\newblock \emph{J. Mach. Learn. Res.}, 3:\penalty0 507--554, 2002.
\newblock URL \url{http://jmlr.org/papers/v3/chickering02b.html}.

\bibitem[Devlin et~al.(2019)Devlin, Chang, Lee, and
  Toutanova]{devlin-etal-2019-bert}
Jacob Devlin, Ming-Wei Chang, Kenton Lee, and Kristina Toutanova.
\newblock {BERT}: {P}re-training of deep bidirectional transformers for
  language understanding.
\newblock In \emph{Proceedings of the 2019 Conference of the North {A}merican
  Chapter of the Association for Computational Linguistics: Human Language
  Technologies, Volume 1 (Long and Short Papers)}, pp.\  4171--4186,
  Minneapolis, Minnesota, June 2019. Association for Computational Linguistics.
\newblock \doi{10.18653/v1/N19-1423}.
\newblock URL \url{https://aclanthology.org/N19-1423}.

\bibitem[Glymour et~al.(2019)Glymour, Zhang, and Spirtes]{clark2019review}
Clark Glymour, Kun Zhang, and Peter Spirtes.
\newblock Review of causal discovery methods based on graphical models.
\newblock \emph{Frontiers in Genetics}, 10:\penalty0 524, 2019.
\newblock ISSN 1664-8021.
\newblock \doi{10.3389/fgene.2019.00524}.
\newblock URL
  \url{https://www.frontiersin.org/article/10.3389/fgene.2019.00524}.

\bibitem[Glymour et~al.(2016)Glymour, Pearl, and Jewell]{glymour2016causal}
Madelyn Glymour, Judea Pearl, and Nicholas~P Jewell.
\newblock \emph{Causal inference in statistics: {A} primer}.
\newblock John Wiley and Sons, 2016.

\bibitem[Gordon et~al.(2012)Gordon, Kozareva, and
  Roemmele]{gordon-etal-2012-semeval}
Andrew Gordon, Zornitsa Kozareva, and Melissa Roemmele.
\newblock {S}em{E}val-2012 task 7: {C}hoice of plausible alternatives: An
  evaluation of commonsense causal reasoning.
\newblock In \emph{*{SEM} 2012: The First Joint Conference on Lexical and
  Computational Semantics {--} Volume 1: Proceedings of the main conference and
  the shared task, and Volume 2: Proceedings of the Sixth International
  Workshop on Semantic Evaluation ({S}em{E}val 2012)}, pp.\  394--398,
  Montr{\'e}al, Canada, 7-8 June 2012. Association for Computational
  Linguistics.
\newblock URL \url{https://aclanthology.org/S12-1052}.

\bibitem[He et~al.(2021)He, Liu, Gao, and Chen]{he2021deberta}
Pengcheng He, Xiaodong Liu, Jianfeng Gao, and Weizhu Chen.
\newblock Deberta: {D}ecoding-enhanced {Bert} with disentangled attention.
\newblock In \emph{9th International Conference on Learning Representations,
  {ICLR} 2021, Virtual Event, Austria, May 3-7, 2021}. OpenReview.net, 2021.
\newblock URL \url{https://openreview.net/forum?id=XPZIaotutsD}.

\bibitem[Hoyer et~al.(2008)Hoyer, Janzing, Mooij, Peters, and
  Sch{\"{o}}lkopf]{hoyer2008nonlinear}
Patrik~O. Hoyer, Dominik Janzing, Joris~M. Mooij, Jonas Peters, and Bernhard
  Sch{\"{o}}lkopf.
\newblock Nonlinear causal discovery with additive noise models.
\newblock In Daphne Koller, Dale Schuurmans, Yoshua Bengio, and L{\'{e}}on
  Bottou (eds.), \emph{Advances in Neural Information Processing Systems 21,
  Proceedings of the Twenty-Second Annual Conference on Neural Information
  Processing Systems, Vancouver, British Columbia, Canada, December 8-11,
  2008}, pp.\  689--696. Curran Associates, Inc., 2008.
\newblock URL
  \url{https://proceedings.neurips.cc/paper/2008/hash/f7664060cc52bc6f3d620bcedc94a4b6-Abstract.html}.

\bibitem[Jin et~al.(2020)Jin, Jin, Zhou, and Szolovits]{jin2020bert}
Di~Jin, Zhijing Jin, Joey~Tianyi Zhou, and Peter Szolovits.
\newblock Is {BERT} really robust? {A} strong baseline for natural language
  attack on text classification and entailment.
\newblock In \emph{The Thirty-Fourth {AAAI} Conference on Artificial
  Intelligence, {AAAI} 2020, The Thirty-Second Innovative Applications of
  Artificial Intelligence Conference, {IAAI} 2020, The Tenth {AAAI} Symposium
  on Educational Advances in Artificial Intelligence, {EAAI} 2020, New York,
  NY, USA, February 7-12, 2020}, pp.\  8018--8025. {AAAI} Press, 2020.
\newblock URL \url{https://aaai.org/ojs/index.php/AAAI/article/view/6311}.

\bibitem[Jin et~al.(2022)Jin, Lalwani, Vaidhya, Shen, Ding, Lyu, Sachan,
  Mihalcea, and Sch{\"{o}}lkopf]{jin-etal-2022-logical}
Zhijing Jin, Abhinav Lalwani, Tejas Vaidhya, Xiaoyu Shen, Yiwen Ding, Zhiheng
  Lyu, Mrinmaya Sachan, Rada Mihalcea, and Bernhard Sch{\"{o}}lkopf.
\newblock Logical fallacy detection.
\newblock In \emph{Findings of the Association for Computational Linguistics:
  EMNLP 2022}, pp.\  7180Ã¢â‚¬â€œ--7198, Abu Dhabi, United Arab
  Emirates, December 2022. Association for Computational Linguistics.
\newblock URL \url{https://arxiv.org/abs/2202.13758}.

\bibitem[K{\i}c{\i}man et~al.(2023)K{\i}c{\i}man, Ness, Sharma, and
  Tan]{kiciman2023causal}
Emre K{\i}c{\i}man, Robert Ness, Amit Sharma, and Chenhao Tan.
\newblock Causal reasoning and large language models: Opening a new frontier
  for causality.
\newblock \emph{arXiv preprint arXiv:2305.00050}, 2023.

\bibitem[Lewis et~al.(2020)Lewis, Liu, Goyal, Ghazvininejad, Mohamed, Levy,
  Stoyanov, and Zettlemoyer]{lewis-etal-2020-bart}
Mike Lewis, Yinhan Liu, Naman Goyal, Marjan Ghazvininejad, Abdelrahman Mohamed,
  Omer Levy, Veselin Stoyanov, and Luke Zettlemoyer.
\newblock {BART}: {D}enoising sequence-to-sequence pre-training for natural
  language generation, translation, and comprehension.
\newblock In \emph{Proceedings of the 58th Annual Meeting of the Association
  for Computational Linguistics}, pp.\  7871--7880, Online, July 2020.
  Association for Computational Linguistics.
\newblock \doi{10.18653/v1/2020.acl-main.703}.
\newblock URL \url{https://aclanthology.org/2020.acl-main.703}.

\bibitem[Liu et~al.(2019)Liu, Ott, Goyal, Du, Joshi, Chen, Levy, Lewis,
  Zettlemoyer, and Stoyanov]{liu2019roberta}
Yinhan Liu, Myle Ott, Naman Goyal, Jingfei Du, Mandar Joshi, Danqi Chen, Omer
  Levy, Mike Lewis, Luke Zettlemoyer, and Veselin Stoyanov.
\newblock {RoBERTa}: {A} robustly optimized {BERT} pretraining approach.
\newblock \emph{CoRR}, abs/1907.11692, 2019.
\newblock URL \url{http://arxiv.org/abs/1907.11692}.

\bibitem[MacCartney \& Manning(2008)MacCartney and
  Manning]{maccartney-manning-2008-modeling}
Bill MacCartney and Christopher~D. Manning.
\newblock Modeling semantic containment and exclusion in natural language
  inference.
\newblock In \emph{Proceedings of the 22nd International Conference on
  Computational Linguistics (Coling 2008)}, pp.\  521--528, Manchester, UK,
  August 2008. Coling 2008 Organizing Committee.
\newblock URL \url{https://aclanthology.org/C08-1066}.

\bibitem[McKay \& Piperno(2014)McKay and Piperno]{mckay2014practical}
Brendan~D. McKay and Adolfo Piperno.
\newblock Practical graph isomorphism, {II}.
\newblock \emph{J. Symb. Comput.}, 60:\penalty0 94--112, 2014.
\newblock \doi{10.1016/j.jsc.2013.09.003}.
\newblock URL \url{https://doi.org/10.1016/j.jsc.2013.09.003}.

\bibitem[Morris et~al.(2020)Morris, Lifland, Yoo, Grigsby, Jin, and
  Qi]{morris-etal-2020-textattack}
John Morris, Eli Lifland, Jin~Yong Yoo, Jake Grigsby, Di~Jin, and Yanjun Qi.
\newblock {T}ext{A}ttack: {A} framework for adversarial attacks, data
  augmentation, and adversarial training in {NLP}.
\newblock In \emph{Proceedings of the 2020 Conference on Empirical Methods in
  Natural Language Processing: System Demonstrations}, pp.\  119--126, Online,
  October 2020. Association for Computational Linguistics.
\newblock \doi{10.18653/v1/2020.emnlp-demos.16}.
\newblock URL \url{https://aclanthology.org/2020.emnlp-demos.16}.

\bibitem[OpenAI(2023)]{openai2023gpt4}
OpenAI.
\newblock {GPT-4} technical report.
\newblock \emph{CoRR}, abs/2303.08774, 2023.
\newblock \doi{10.48550/arXiv.2303.08774}.
\newblock URL \url{https://doi.org/10.48550/arXiv.2303.08774}.

\bibitem[Ouyang et~al.(2022)Ouyang, Wu, Jiang, Almeida, Wainwright, Mishkin,
  Zhang, Agarwal, Slama, Ray, Schulman, Hilton, Kelton, Miller, Simens, Askell,
  Welinder, Christiano, Leike, and Lowe]{ouyang2022instructGPT}
Long Ouyang, Jeff Wu, Xu~Jiang, Diogo Almeida, Carroll~L. Wainwright, Pamela
  Mishkin, Chong Zhang, Sandhini Agarwal, Katarina Slama, Alex Ray, John
  Schulman, Jacob Hilton, Fraser Kelton, Luke Miller, Maddie Simens, Amanda
  Askell, Peter Welinder, Paul~F. Christiano, Jan Leike, and Ryan Lowe.
\newblock Training language models to follow instructions with human feedback.
\newblock \emph{CoRR}, abs/2203.02155, 2022.
\newblock \doi{10.48550/arXiv.2203.02155}.
\newblock URL \url{https://doi.org/10.48550/arXiv.2203.02155}.

\bibitem[Pearl(1988)]{pearl1988probabilistic}
Judea Pearl.
\newblock \emph{Probabilistic reasoning in intelligent systems: {N}etworks of
  plausible inference}.
\newblock Morgan Kaufmann, 1988.

\bibitem[Pearl(2009)]{pearl2009causality}
Judea Pearl.
\newblock \emph{Causality: {M}odels, reasoning and inference (2nd ed.)}.
\newblock Cambridge University Press, 2009.

\bibitem[Peters et~al.(2017)Peters, Janzing, and
  Sch{\"o}lkopf]{peters2017elements}
Jonas Peters, Dominik Janzing, and Bernhard Sch{\"o}lkopf.
\newblock \emph{Elements of causal inference: {F}oundations and learning
  algorithms}.
\newblock The MIT Press, 2017.
\newblock URL \url{https://mitpress.mit.edu/books/elements-causal-inference}.

\bibitem[Qin et~al.(2019)Qin, Bosselut, Holtzman, Bhagavatula, Clark, and
  Choi]{qin-etal-2019-counterfactual}
Lianhui Qin, Antoine Bosselut, Ari Holtzman, Chandra Bhagavatula, Elizabeth
  Clark, and Yejin Choi.
\newblock Counterfactual story reasoning and generation.
\newblock In \emph{Proceedings of the 2019 Conference on Empirical Methods in
  Natural Language Processing and the 9th International Joint Conference on
  Natural Language Processing (EMNLP-IJCNLP)}, pp.\  5043--5053, Hong Kong,
  China, November 2019. Association for Computational Linguistics.
\newblock \doi{10.18653/v1/D19-1509}.
\newblock URL \url{https://aclanthology.org/D19-1509}.

\bibitem[Radford et~al.(2019)Radford, Wu, Child, Luan, Amodei, and
  Sutskever]{radford2019language}
Alec Radford, Jeffrey Wu, Rewon Child, David Luan, Dario Amodei, and Ilya
  Sutskever.
\newblock Language models are unsupervised multitask learners.
\newblock \emph{OpenAI Blog}, 1\penalty0 (8), 2019.

\bibitem[Sanh et~al.(2019)Sanh, Debut, Chaumond, and Wolf]{sanh2019distilbert}
Victor Sanh, Lysandre Debut, Julien Chaumond, and Thomas Wolf.
\newblock {DistilBERT}, a distilled version of {BERT:} {S}maller, faster,
  cheaper and lighter.
\newblock \emph{CoRR}, abs/1910.01108, 2019.
\newblock URL \url{http://arxiv.org/abs/1910.01108}.

\bibitem[Sap et~al.(2019{\natexlab{a}})Sap, Bras, Allaway, Bhagavatula, Lourie,
  Rashkin, Roof, Smith, and Choi]{sap2019atomic}
Maarten Sap, Ronan~Le Bras, Emily Allaway, Chandra Bhagavatula, Nicholas
  Lourie, Hannah Rashkin, Brendan Roof, Noah~A. Smith, and Yejin Choi.
\newblock {ATOMIC:} an atlas of machine commonsense for if-then reasoning.
\newblock In \emph{The Thirty-Third {AAAI} Conference on Artificial
  Intelligence, {AAAI} 2019, The Thirty-First Innovative Applications of
  Artificial Intelligence Conference, {IAAI} 2019, The Ninth {AAAI} Symposium
  on Educational Advances in Artificial Intelligence, {EAAI} 2019, Honolulu,
  Hawaii, USA, January 27 - February 1, 2019}, pp.\  3027--3035. {AAAI} Press,
  2019{\natexlab{a}}.
\newblock \doi{10.1609/aaai.v33i01.33013027}.
\newblock URL \url{https://doi.org/10.1609/aaai.v33i01.33013027}.

\bibitem[Sap et~al.(2019{\natexlab{b}})Sap, Rashkin, Chen, {Le Bras}, and
  Choi]{Sap2019SocialIQA}
Maarten Sap, Hannah Rashkin, Derek Chen, Ronan {Le Bras}, and Yejin Choi.
\newblock Social iqa: {C}ommonsense reasoning about social interactions.
\newblock In \emph{EMNLP 2019}, 2019{\natexlab{b}}.

\bibitem[Shimizu et~al.(2006)Shimizu, Hoyer, Hyv{\"{a}}rinen, and
  Kerminen]{shimizu2006linear}
Shohei Shimizu, Patrik~O. Hoyer, Aapo Hyv{\"{a}}rinen, and Antti~J. Kerminen.
\newblock A linear non-gaussian acyclic model for causal discovery.
\newblock \emph{J. Mach. Learn. Res.}, 7:\penalty0 2003--2030, 2006.
\newblock URL \url{http://jmlr.org/papers/v7/shimizu06a.html}.

\bibitem[Shleifer \& Rush(2020)Shleifer and Rush]{shleifer2020pretrained}
Sam Shleifer and Alexander~M. Rush.
\newblock Pre-trained summarization distillation.
\newblock \emph{CoRR}, abs/2010.13002, 2020.
\newblock URL \url{https://arxiv.org/abs/2010.13002}.

\bibitem[Spirtes \& Zhang(2016)Spirtes and Zhang]{spirtes2016causal}
Peter Spirtes and Kun Zhang.
\newblock Causal discovery and inference: {C}oncepts and recent methodological
  advances.
\newblock In \emph{Applied informatics}, volume~3, pp.\  1--28. SpringerOpen,
  2016.

\bibitem[Spirtes et~al.(1993)Spirtes, Glymour, and
  Scheines]{spirtes1993causation}
Peter Spirtes, Clark Glymour, and Richard Scheines.
\newblock Causation, prediction, and search.
\newblock 1993.

\bibitem[Spirtes et~al.(2000)Spirtes, Glymour, and
  Scheines]{spirtes2000causation}
Peter Spirtes, Clark Glymour, and Richard Scheines.
\newblock \emph{Causation, Prediction, and Search, Second Edition}.
\newblock Adaptive computation and machine learning. {MIT} Press, 2000.
\newblock ISBN 978-0-262-19440-2.

\bibitem[Taori et~al.(2023)Taori, Gulrajani, Zhang, Dubois, Li, Guestrin,
  Liang, and Hashimoto]{taori2023alpaca}
Rohan Taori, Ishaan Gulrajani, Tianyi Zhang, Yann Dubois, Xuechen Li, Carlos
  Guestrin, Percy Liang, and Tatsunori~B. Hashimoto.
\newblock Stanford alpaca: {A}n instruction-following llama model.
\newblock \url{https://github.com/tatsu-lab/stanford\_alpaca}, 2023.

\bibitem[Touvron et~al.(2023)Touvron, Lavril, Izacard, Martinet, Lachaux,
  Lacroix, Rozi{\`{e}}re, Goyal, Hambro, Azhar, Rodriguez, Joulin, Grave, and
  Lample]{touvron2023llama}
Hugo Touvron, Thibaut Lavril, Gautier Izacard, Xavier Martinet, Marie{-}Anne
  Lachaux, Timoth{\'{e}}e Lacroix, Baptiste Rozi{\`{e}}re, Naman Goyal, Eric
  Hambro, Faisal Azhar, Aur{\'{e}}lien Rodriguez, Armand Joulin, Edouard Grave,
  and Guillaume Lample.
\newblock Llama: {O}pen and efficient foundation language models.
\newblock \emph{CoRR}, abs/2302.13971, 2023.
\newblock \doi{10.48550/arXiv.2302.13971}.
\newblock URL \url{https://doi.org/10.48550/arXiv.2302.13971}.

\bibitem[Tu et~al.(2023)Tu, Ma, and Zhang]{tu2023causal}
Ruibo Tu, Chao Ma, and Cheng Zhang.
\newblock Causal-discovery performance of chatgpt in the context of neuropathic
  pain diagnosis.
\newblock \emph{arXiv preprint arXiv:2301.13819}, 2023.

\bibitem[Willig et~al.(2023)Willig, Ze{\v{c}}evi{\'c}, Dhami, and
  Kersting]{willig2023probing}
Moritz Willig, Matej Ze{\v{c}}evi{\'c}, Devendra~Singh Dhami, and Kristian
  Kersting.
\newblock Probing for correlations of causal facts: Large language models and
  causality, 2023.
\newblock URL \url{https://openreview.net/forum?id=UPwzqPOs4-}.

\bibitem[Wolf et~al.(2020)Wolf, Debut, Sanh, Chaumond, Delangue, Moi, Cistac,
  Rault, Louf, Funtowicz, Davison, Shleifer, von Platen, Ma, Jernite, Plu, Xu,
  Le~Scao, Gugger, Drame, Lhoest, and Rush]{wolf2019transformers}
Thomas Wolf, Lysandre Debut, Victor Sanh, Julien Chaumond, Clement Delangue,
  Anthony Moi, Pierric Cistac, Tim Rault, Remi Louf, Morgan Funtowicz, Joe
  Davison, Sam Shleifer, Patrick von Platen, Clara Ma, Yacine Jernite, Julien
  Plu, Canwen Xu, Teven Le~Scao, Sylvain Gugger, Mariama Drame, Quentin Lhoest,
  and Alexander Rush.
\newblock Transformers: {S}tate-of-the-art natural language processing.
\newblock In \emph{Proceedings of the 2020 Conference on Empirical Methods in
  Natural Language Processing: System Demonstrations}, pp.\  38--45, Online,
  October 2020. Association for Computational Linguistics.
\newblock \doi{10.18653/v1/2020.emnlp-demos.6}.
\newblock URL \url{https://aclanthology.org/2020.emnlp-demos.6}.

\bibitem[Xie et~al.(2023)Xie, Li, and Kan]{xie2023echo}
Yuxi Xie, Guanzhen Li, and Min-Yen Kan.
\newblock Echo: Event causality inference via human-centric reasoning.
\newblock \emph{arXiv preprint arXiv:2305.14740}, 2023.

\bibitem[Ze{\v{c}}evi{\'c} et~al.(2023)Ze{\v{c}}evi{\'c}, Willig, Dhami, and
  Kersting]{zevcevic2023causal}
Matej Ze{\v{c}}evi{\'c}, Moritz Willig, Devendra~Singh Dhami, and Kristian
  Kersting.
\newblock Causal parrots: Large language models may talk causality but are not
  causal.
\newblock \emph{arXiv preprint arXiv:2308.13067}, 2023.

\bibitem[Zhang \& Hyv{\"a}rinen(2009)Zhang and
  Hyv{\"a}rinen]{zhang2009causality}
Kun Zhang and Aapo Hyv{\"a}rinen.
\newblock Causality discovery with additive disturbances: {A}n
  information-theoretical perspective.
\newblock In \emph{Machine Learning and Knowledge Discovery in Databases:
  European Conference, ECML PKDD 2009, Bled, Slovenia, September 7-11, 2009,
  Proceedings, Part II 20}, pp.\  570--585. Springer, 2009.

\bibitem[Zhang et~al.(2020)Zhang, Lyu, and
  Callison-Burch]{zhang-etal-2020-reasoning}
Li~Zhang, Qing Lyu, and Chris Callison-Burch.
\newblock Reasoning about goals, steps, and temporal ordering with {W}iki{H}ow.
\newblock In \emph{Proceedings of the 2020 Conference on Empirical Methods in
  Natural Language Processing (EMNLP)}, pp.\  4630--4639, Online, November
  2020. Association for Computational Linguistics.
\newblock \doi{10.18653/v1/2020.emnlp-main.374}.
\newblock URL \url{https://aclanthology.org/2020.emnlp-main.374}.

\bibitem[Zhang et~al.(2022)Zhang, Roller, Goyal, Artetxe, Chen, Chen, Dewan,
  Diab, Li, Lin, Mihaylov, Ott, Shleifer, Shuster, Simig, Koura, Sridhar, Wang,
  and Zettlemoyer]{zhang2022opt}
Susan Zhang, Stephen Roller, Naman Goyal, Mikel Artetxe, Moya Chen, Shuohui
  Chen, Christopher Dewan, Mona~T. Diab, Xian Li, Xi~Victoria Lin, Todor
  Mihaylov, Myle Ott, Sam Shleifer, Kurt Shuster, Daniel Simig, Punit~Singh
  Koura, Anjali Sridhar, Tianlu Wang, and Luke Zettlemoyer.
\newblock {OPT:} open pre-trained transformer language models.
\newblock \emph{CoRR}, abs/2205.01068, 2022.
\newblock \doi{10.48550/arXiv.2205.01068}.
\newblock URL \url{https://doi.org/10.48550/arXiv.2205.01068}.

\end{thebibliography}
\bibliographystyle{iclr2024_conference}

\newpage
\appendix

\section{Implementation Details}\label{appd:implementation}
When finetuning on our data,  for GPT-based models, we use the default settings of the OpenAI finetuning API; and for BERT-based models, 
we use the \texttt{transformers} library \citep{wolf2019transformers} and train the models on a server with an NVIDIA Tesla A100 GPU with 40G of memory. To fit for the GPU memory, we set the batch size to be 8. We use the validation set to tune the learning rate, which takes value in \{2e-6,  5e-6, 1e-5, 2e-5, 5e-5\}; dropout rate, which takes value in \{0, 0.1, 0.2, 0.3\}; and weight decay, which takes value in \{1e-4, 1e-5\}. We train the models until convergence, which is usually around ten epochs.

\paragraph{Prompts}
When querying the autoregressive LLMs, we formulate the prompt as follows:

\textit{Question: \texttt{[premise]}}

\textit{Can we deduct the following: \texttt{[hypothesis]}? Just answer "Yes" or "No."}

\textit{Answer:}

\section{Generating Natural Stories}\label{appd:story}

To generate the natural stories based on our symbolic expressions, we utilize the state-of-the-art LLM, GPT-4, which is very good at story generation. We design detailed instructions in the prompt, and generate around 200 stories in our case study. We show two examples stories in \cref{tab:story}, and the report the overall statistics in \cref{tab:story_stats}.
\begin{table}[ht]
    \centering
    \begin{tabular}{p{1cm}p{6cm}p{6cm}l}
    \toprule
& Example 1 (Label=Negative) & Example 2 (Label=Positive) \\ \midrule
Symbolic Form        
& \textit{Premise:} Suppose there is a closed system of 2 variables, A and B. All the statistical relations among these 2 variables are as follows: A correlates with B. 
\newline
\textit{Hypothesis:} A directly affects B.	
\newline
\textit{Relation between the promise and hypothesis:} The premise does not a necessary condition for the hypothesis.
& \textit{Premise:} Suppose there is a closed system of 3 variables, A, B and C. All the statistical relations among these 3 variables are as follows: A correlates with C. B correlates with C. However, A is independent of B. 
\newline
\textit{Hypothesis:} A directly affects C.
\newline
\textit{Relation between the promise and hypothesis: }The premise is a necessary condition for the hypothesis. So if the premise is true, the hypothesis must be true. \\ \hline
Natural Story         
& \textit{Premise:} Suppose there is a closed system of 2 variables, ice cream sales and swimming pool attendance. All the statistical relations among these 2 variables are as follows: ice cream sales correlate with swimming pool attendance.
\newline
\textit{Hypothesis:} Ice cream sales directly affect swimming pool attendance.
\newline
\textit{Relation between the premise and hypothesis:} The premise does not provide a necessary condition for the hypothesis. The correlation between ice cream sales and swimming pool attendance could be due to a third variable, such as hot weather, which increases both ice cream sales and swimming pool attendance. Therefore, it is not necessarily true that ice cream sales directly affect swimming pool attendance.
& 
\textit{Premise:} Let's consider three factors: eating junk food (A), obesity (C), and watching television (B). There is a correlation between eating junk food and obesity, and between watching television and obesity. However, eating junk food and watching television are independent from each other.
\newline
\textit{Hypothesis:} Eating junk food directly affects obesity.
\newline
\textit{Relation between the premise and hypothesis:} The premise provides the necessary conditions for the hypothesis. It establishes the independent variables A (eating junk food) and B (watching television) and their correlations with obesity. Given that these are true, it supports the hypothesis that eating junk food directly affects obesity.
\\
         \bottomrule
    \end{tabular}
    \caption{Examples of natural stories generated based on the symbolic form in our \ourdata dataset, showing the broad application value of our dataset as the starting point for various verbalizations of the correlation-to-causation inference task.}
    \label{tab:story}
\end{table}
\begin{table}[h]
    \centering\small
    \begin{tabular}{lcccccc}
    \toprule
Test Set Size  & 102 
\\
Dev Set Size & 102 
\\
\# Tokens/Premise & 64.88 
\\
\# Tokens/Hypothesis & 13.54 
\\
\# Tokens/Explanation & 64.66 
\\
\% Positive Labels & 1.67 
\\
\bottomrule
    \end{tabular}
    \caption{Statistics of our generated natural stories. We report the number of samples in the test and development sets; number of tokens per premise (\# Tokens/Premise), hypothesis (\# Tokens/Hypothesis), and explanation (\# Tokens/Explanation); and percentage of the positive labels (\% Positive Labels).}\label{tab:story_stats}
\end{table}

For more information, the exact prompt we use is ``\textit{Here is a causal inference rule: \texttt{[symbolic form]} Please provide a real-world example instantiating this phenomenon. Format it also as "Premise:", "Hypothesis:", and "Relation between the promise and hypothesis:".}''

\section{Templates and Paraphrases}\label{appd:templates}
We use the verbalization templates in \cref{tab:paraphrase} to compose the hypotheses for all six causal relations.
\begin{table}[ht]
    \centering \small
    \begin{tabular}{ll}
\toprule
\textbf{Causal Relation} & \textbf{Hypothesis Template} \\\midrule
Is-Parent & \texttt{\{Var i\}} directly causes \texttt{\{Var j\}}. \\
Is-Ancestor & \texttt{\{Var i\}} causes something else which causes \texttt{\{Var j\}}. \\
Is-Child & \texttt{\{Var j\}} directly causes \texttt{\{Var i\}}. \\
Is-Descendant & \texttt{\{Var j\}} is a cause for \texttt{\{Var i\}}, but not a direct one. \\
Has-Collider & There exists at least one collider (i.e., common effect) of \texttt{\{Var i\}} and \texttt{\{Var j\}}. \\
Has-Confounder & There exists at least one confounder (i.e., common cause) of \texttt{\{Var i\}} and \texttt{\{Var j\}}. \\
\midrule
\textbf{\textit{Paraphrases}} \\

Is-Parent & \texttt{\{Var i\}} directly affects \texttt{\{Var j\}}. \\
Is-Ancestor & \texttt{\{Var i\}} influences \texttt{\{Var j\}} through some mediator(s). \\
Is-Child & \texttt{\{Var j\}} directly affects \texttt{\{Var i\}}. \\
Is-Descendant & \texttt{\{Var j\}} influences \texttt{\{Var i\}} through some mediator(s). \\
Has-Collider & \texttt{\{Var i\}} and \texttt{\{Var j\}} together cause some other variable(s). \\
Has-Confounder & Some variable(s) cause(s) both \texttt{\{Var i\}} and \texttt{\{Var j\}}. \\
\bottomrule
    \end{tabular}
    \caption{Templates and their paraphrases for each causal relation in the hypothesis. We use \texttt{\{Var i\}} and \texttt{\{Var j\}} as placeholders for the two variables.}
    \label{tab:paraphrase}
\end{table}

\section{Change Log for the Dataset Version Update}\label{appd:version}

\begin{table}[ht]
    \centering \small
    \begin{tabular}{lp{3.3cm}llll}
    \toprule
    Two Equivalent Forms & Duplication Property   &  De-Duplication Method \\ \midrule
\tworowbrace { }     Is-Parent(\texttt{i}, \texttt{j}) & \multirow{2}{*}{Two exact same strings} & \multirow{2}{*}{Keep only one, by forcing $i<j$} \\
\vspace{0.4em}
{ }{ } Is-Child(\texttt{j}, \texttt{i})\\
 
\tworowbrace { }     Is-Ancestor(\texttt{i}, \texttt{j}) (Original)  & Two different strings, but  & \multirow{2}{*}{Randomly sample one out of the two} \\
\vspace{0.4em}
{ }{ } Is-Descendent(\texttt{j}, \texttt{i}) (Original)  &  semantically equivalent    \\
 
\tworowbrace { }         Is-Ancestor(\texttt{i}, \texttt{j}) (Paraphrased) & \multirow{2}{*}{Two exact same strings} & \multirow{2}{*}{Keep only one, by forcing $i<j$}\\
\vspace{0.4em}
{ }{ } Is-Descendent(\texttt{j}, \texttt{i}) (Paraphrased) \\

\tworowbrace { }     Has-Collider(\texttt{i}, \texttt{j})  & Two different strings, but  & \multirow{2}{*}{Randomly sample one out of the two} \\
\vspace{0.4em}
{ }{ } Has-Collider(\texttt{j}, \texttt{i})  &  semantically equivalent    \\
 
\tworowbrace { }     Has-Confounder(\texttt{i}, \texttt{j})  & Two different strings, but  & \multirow{2}{*}{Randomly sample one out of the two} \\
\vspace{0.4em}
{ }{ } Has-Confounder(\texttt{j}, \texttt{i})  &  semantically equivalent    \\
    \bottomrule
    \end{tabular}
    
    \caption{De-duplication methods for the six causal relation types and their verbalizations.}
    \label{tab:version}
\end{table}
\paragraph{De-Duplication Strategy}
As mentioned in \cref{sec:stats} in the main paper, our original dataset (v1.0) has duplication due to symmetric relations and verbalizations. 
We introduce in \cref{tab:version} several reasons for why duplicated hypotheses exist in our original data. One typical reason is symmetric relations such as Is-Parent(A, B) and Is-Child(B, A), and, similarly, the paraphrased version of Is-Ancestor(A, B) and Is-Descendent(B, A).
Another typical reason is the semantic equivalence in the verbalization templates, which applies to the Has-Collider and Has-Confounder relations. For example, the verbalized texts of Has-Collider(A, B) and Collider(B, A) are ``There exists at least one collider (i.e., common effect) of \{A and B, B and A\},'' respectively, which are semantically-equivalent paraphrases of each other, so we randomly keep one out of the two.

\paragraph{Resulting Dataset Statistics after De-Duplication}

Since the reason for duplication in the first place is due to symmetry in the causal relation, or verbalization, the resulting new data, \ourdata v2.0, is exactly a half of the original data. As we reported previously in \cref{tab:stats} of \cref{sec:stats}, the total number of samples cuts down to half, while the label distribution and all other properties are the same. To compose each split, we apply the same de-duplication method for the test, train, and development sets. We notice that some duplicates are across the splits, so we prioritize keeping the test and training sets untouched (to minimally affect the experimental results), and then reduce the development set by removing the cross-split duplicates, namely:
\begin{itemize}
    \item test\_2.0 = deduplicate(test\_1.0)
    \item train\_2.0 = deduplicate(train\_1.0)
    \item dev\_2.0 = deduplicate(dev\_1.0) $\backslash$ \{test\_2.0, train\_2.0\}
\end{itemize}
We expect minimal or almost no change to the experimental results. In case of the slight possibility that this change in the development set might affect the model selection in the training process, future work can feel free to re-train the models and update the exact performance number.

\section{Spurious Correlation Analysis}\label{appd:spurious}
The inspirations of our two robustness tests (paraphrasing and variable refactorization) come from our data analysis. 
We check for spurious correlations in the data by reporting in \cref{tab:ngram} the point-wise mutual information (PMI) between the label and any n-gram with no more than four tokens.
In addition, we also report the difference of the PMI with the two labels in the $|\text{Diff}|$ column of \cref{tab:ngram}, and report the top 10 n-grams.

The design spirit for our robustness test is that if
the models' correct judgment relies on exploiting these spurious correlations, then such reliance will be broken in our perturbations.

\begin{table}[ht]
    \centering \small
\begin{tabular}{lcccc}
\toprule
N-Gram &  PMI w/ Non-Ent. Label &  PMI w/ Ent. Label &    $|\text{Diff}|$ \\
\midrule
a cause            &  1.692209 & -1.025611 &  2.717820 \\
a cause for        &  1.663640 & -0.983790 &  2.647430 \\
A causes           &  1.640679 & -0.951610 &  2.592289 \\
A causes something &  1.621820 & -0.926075 &  2.547895 \\
a direct           &  1.606052 & -0.905316 &  2.511369 \\
a direct one       &  1.592673 & -0.888107 &  2.480781 \\
for D              &  1.584826 & -0.878180 &  2.463006 \\
for D but          &  1.583897 & -0.877014 &  2.460911 \\
for E              &  1.582980 & -0.875864 &  2.458844 \\
for E but          &  1.582074 & -0.874728 &  2.456802 \\

\bottomrule
\end{tabular}

    \caption{PMI between the labels and n-grams. The labels include non-entailment (Non-Ent.) and entailment (Ent.). And the n-grams include all with no more than four words. The $|\text{Diff}|$ column shows the absolute value of the difference between the PMIs with two labels.
    We show the top 10 n-grams with the largest differences of their PMIs with the two classes in the $|\text{Diff}|$ column.
    }
    \label{tab:ngram}
\end{table}

We can see that some spurious correlations are rooted in the framing of the hypothesis, such as ``a cause (for)'', and ``a direct (one)'' (which we use the paraphrasing task to break), and others are connected to the variable names, such as ``for D (but)'' and ``for E (but)'' (which we use the variable refactorization to break).

\section{Fine-Grained Error Analysis}\label{appd:error}
In addition to the fine-grained analysis by causal relation type in \cref{tab:finetune_by_class} for fine-tuned models, we also report such error analysis for non-finetuned models in \cref{tab:by_class}.
\begin{table}[ht]
\centering \small
\begin{tabular}{llcccccccc}
\toprule
{Selected Models}      & {Relation Type} & {F1} & {Precision} & {Recall} & {Accuracy} \\ \midrule
GPT-3.5             & All                    & 21.69       & 17.79      & 27.78      & 69.46        \\
GPT-3.5             & Is-Parent              & 8.82        & 100        & 4.62       & 83.47        \\
GPT-3.5             & Is-Ancestor            & 0           & 0          & 0          & 90.67        \\
GPT-3.5             & Is-Child               & 9.84        & 100        & 5.17       & 85.33        \\
GPT-3.5             & Is-Descendant          & 14.29       & 11.9       & 17.86      & 84           \\
GPT-3.5             & Has-Collider           & 34.24       & 25.51      & 52.07      & 35.12        \\
GPT-3.5             & Has-Confounder         & 15.33       & 8.86       & 56.76      & 37.8         \\ \hline
GPT-4               & All                    & 29.08       & 20.92      & 47.66      & 64.6         \\
GPT-4               & Is-Parent              & 0           & 0          & 0          & 82.67        \\
GPT-4               & Is-Ancestor            & 30.77       & 31.25      & 30.3       & 88           \\
GPT-4               & Is-Child               & 0           & 0          & 0          & 84.53        \\
GPT-4               & Is-Descendant          & 26.98       & 17.35      & 60.71      & 75.47        \\
GPT-4               & Has-Collider           & 44.1        & 30.18      & 81.82      & 32.71        \\
GPT-4               & Has-Confounder         & 20.67       & 11.53      & 100        & 23.86        \\ \hline
RoBERTa MNLI        & All                    & 22.79       & 34.73      & 16.96      & 82.5         \\
RoBERTa MNLI        & Is-Parent              & 0           & 0          & 0          & 82.67        \\
RoBERTa MNLI        & Is-Ancestor            & 0           & 0          & 0          & 91.2         \\
RoBERTa MNLI        & Is-Child               & 0           & 0          & 0          & 84.53        \\
RoBERTa MNLI        & Is-Descendant          & 0           & 0          & 0          & 92.53        \\
RoBERTa MNLI        & Has-Collider           & 43.45       & 39.73      & 47.93      & 59.52        \\
RoBERTa MNLI        & Has-Confounder         & 0           & 0          & 0          & 84.45        \\ \bottomrule
\end{tabular}
\caption{Fine-grained evaluation results for some selected non-fine-tuned models.}
\label{tab:by_class}
\end{table}

These results are particularly revealing, showing how off-the-shelf models perform in recognizing specific relations. Specifically, GPT-3.5 cannot recognize ancestor relations, whereas GPT-4 fails at all direct causation recognition with parents and children. And RoBERTa MNLI only did collider relation relatively correctly. Note that, when the F1 score is zero, the accuracy number is a result of always predicting the negative class of that relation.

\section{LLM Performance Optimization}
\label{appd:optimization}

Since our experiments in \cref{sec:0shot} are based on plain, zero-shot prompts, we explore whether better prompting strategies could improve the performance. We enhance the query prompt by incorporating several strategies: (1) Utilizing a system prompt that specifies the model's expertise (``You are a highly intelligent question-answering bot with profound knowledge of causal inference.''); (2) Including a pair of few-shot examples, one positive and one negative; (3) Implementing chain-of-thought prompting with ``Let's think step by step.'' to encourage the language model to generate step-by-step reasoning. In \cref{tab:optimize_res}, we present the evaluation results on the relatively affordable model, GPT-3.5, where the optimized prompt leads to a 4-point improvement in F1 over the original performance. However, we can see that despite the deployment of all three strategies, the model continues to struggle with this challenging task.

\begin{table}[h]
    \centering \small
    \begin{tabular}{lccccc}
    \toprule
    & F1 & Precision & Recall
    & Accuracy\\ \midrule
    GPT-3.5 (plain query; original) & 21.69 & 17.79 & 27.78 & 69.46 \\

    GPT-3.5 (enhanced query) & 25.44 & 17.29 & 48.11 & 52.01 \\
    
    \bottomrule
    \end{tabular}
    \caption{Performance of GPT-3.5 with different queries. We quote the original performance from \cref{tab:res}.
    }
    \label{tab:optimize_res}
\vspace{-3mm}
\end{table}

\end{document}